# Convolutional Point-set Representation: A Convolutional Bridge Between a Densely Annotated Image and 3D Face Alignment

Yuhang Wu, Le Anh Vu Ha, Xiang Xu, Ioannis A. Kakadiaris

Computational Biomedicine Lab, University of Houston
{ywu35, hale4, xxu18, ikakadia}@central.uh.edu

**Abstract.** We present a robust method for estimating the facial pose and shape information from a densely annotated facial image. The method relies on Convolutional Point-set Representation (CPR), a carefully designed matrix representation to summarize different layers of information encoded in the set of detected points in the annotated image. The CPR disentangles the dependencies of shape and different pose parameters and enables updating different parameters in a sequential manner via convolutional neural networks and recurrent layers. When updating the pose parameters, we sample reprojection errors along with a predicted direction and update the parameters based on the pattern of reprojection errors. This technique boosts the model's capability in searching a local minimum under challenging scenarios. We also demonstrate that annotation from different sources can be merged under the framework of CPR and contributes to outperforming the current state-of-the-art solutions for 3D face alignment. Experiments indicate the proposed CPRFA (CPR-based Face Alignment) significantly improves 3D alignment accuracy when the densely annotated image contains noise and missing values, which is common under "in-the-wild" acquisition scenarios.

## 1 Introduction

This work aims to improve estimation of 3D pose and shape information from a Densely Annotated Facial Image (DAFI), as part of research in the face analysis domain incorporating the recent developments of image dense annotation approaches (*e.g.*, DensePose[1], MaskRCNN [2]). Solving this problem contributes to the applications that require 3D transformations of a human face, such as facial animation, augmented reality, and 3D-aided face recognition. The proposed method can be generalized to solve other object pose estimation and 3D deformation problems.

The problem of 3D face alignment is to register a generic 3D facial model onto a 2D image through model deformation and pose estimation. The objective is to minimize the reprojection error between the projected model vertices and their corresponding observed positions on the 2D image. To achieve accurate 3D face alignment, most of the current solutions require annotating a sparse set of 3D points (*e.g.*, 68 points) and detect its corresponding positions on the 2D image. However, while the number of accurately annotated 2D landmarks is limited in complex scenarios (*e.g.*, occluded facial image and large head pose variations), limited numbers of correspondences often fail to provide enough regularizations to align a dense facial model. Thanks to the recent development of deep network, especially the fully convolutional network [3] and its extensions (*e.g.*, U-Net [4], Hour-glass Net [5]), a densely annotated facial image (DAFI) is available. DAFI annotate each pixel in the original image with a new label, which can be an index of 3D vertex on a 3D model [6], or a similarity value to a template image [7]. Compared with the annotation generated by automatic landmark detectors, DAFI provides more detailed alignment clues for 3D alignment and potentially improves the alignment accuracy. Following this line of research, Guler *et al.* [7] proposed a quantized regression approach to obtain DAFI by constructing the mapping between image and the UV map of the 3D model. Crispell *et al.* [6] directly estimated the projected normalized coordinate code (PNCC [8]) on the facial image using an U-Net and obtained dense landmarks on the human face. This type of DAFI is employed in this paper and depicted in Fig. 2b. Yu *et al.* [9] constructed the dense facial correspondences between an image and a frontal facial model using auto-encoder through predicting a correspondence flow and a matchability mask. Jackson *et al.* [9] estimated a 3D volumetric model of the facial image through hour-glass networks.

Surprisingly, although well-designed deep networks are trained to estimate the DAFI, to the best of our knowledge, no deep network is designed for distilling 3D knowledge from the DAFI. In [6,9], on-line optimization approaches are employed to estimate the parameters of 3D model. These approaches, as depicted in Fig. 1a, may perform well when the densely annotated image is highly accurate. Our experiments indicate that DAFI contains a large amount of missing and noisy annotations when the image is obtained under "in-the-wild" acquisition conditions. Handling such variations is beyond the capability of the online optimization approaches.

In another line of research, deep networks are employed to directly estimate 3D shape and pose parameters from 2D images without using DAFI (as depicted in Fig. 1b). Zhu *et al.* [8] and Jourabloo *et al.* [10] employed a cascaded chain of deep networks to jointly estimate the shape and pose parameters of 3DMM. Recently, Ja *et al.* [11] and Bhagavatula *et al.* [12] employed a single, well-designed network to accomplish the same task, which appears to be more efficient. However, in these approaches, because all the pose parameters are estimated jointly, the model is unaware of prior dependencies between the parameters. This may cause a problem as any untrained pattern (distribution change) in the image may generate problematic responses on unknown target shape and pose parameters. This is detrimental to minimizing the final reprojection error. This can be partially demonstrated by a recent evaluation of Jourabloo *et al.*[10] under synthesized occlusions [13]. The experiment indicates that [10] is sensitive to untrained occlusions.

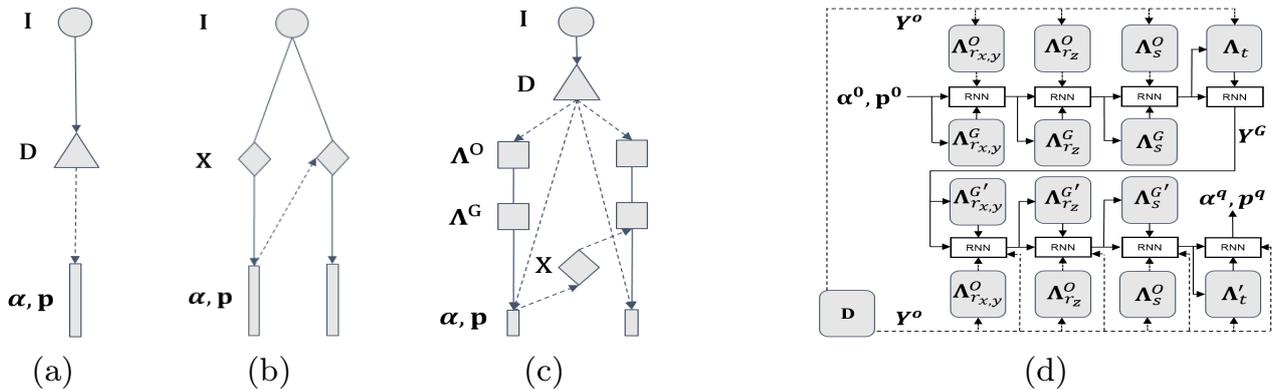

Fig. 1: (a,b,c) Depicted are the different frameworks for 3D face alignment. For (b,c), only the last two cascaded stages are depicted. Arrows with solid lines represent the mappings learned by deep neural networks. (a) Depiction of DAFI-based on-line optimization [6,9]. (b) Depiction of deep learning-based cascaded regression [8,10]. (c) Depiction of the proposed CPR-based consistency-aware learning. (d) Depiction of the two-rounds parameter updating (expanded version of (c)). Different types of CPRs are employed to update the parameters in (d). Arrows with solid lines represent the mappings that allow gradient back-propagation.

In this work, we propose a robust method to distill the reliable knowledge from DAFI for 3D face alignment based on convolutional neural networks. It is constructed based on the insight that DAFI encodes geometry information in the relations among pairs or triplets of annotated pixels. To summarize high-order relations, we represent the point(landmark)-set in DAFI as a 2D matrices, so that recent advances in convolution neural networks can be used to seek a non-linear mapping from the noisy DAFI to the target parameters. As opposed to the previous solutions [6,8–13], our solution disentangles the impact of parameters between pose estimations. Note that a recent research proposed by Bas *et al.*[14] optimizes each pose parameter based on the UV map of the facial image with a spatial transformer network [15]. Different from [14], our model is based on DAFI and demonstrated to be robust and effective with extensive experimental results. To boost the performance of CPR for 3D face alignment, a new approach is proposed to update the model parameters. Thanks to DAFI, reprojection errors can be sampled online along with a predicted direction generated by deep networks. Based on the predicted direction, the step size of parameter updating is determined via observing the distributions of the reprojeciton errors sampled on the path. This approach significantly improves the robustness of model updating under relatively complex environments. Finally, we propose a method to fuse multiple CPRs generated by different image annotation approaches. It allows state-of-the-art landmark detectors [16,17] to contribute to 3D model alignment under the framework of CPR, which provides more reliable clues via localizing fiducial points on the face.

In summary, the proposed work makes the following contributions: (i) to the best of our knowledge, this is the first effort to distill the high-level knowl-edge from DAFI through convolutional neural network based on the transformation order of the weak-perspective projection model; (ii) we propose a new consistency-aware parameter updating model that samples the reprojection along a predicted direction on-line; (iii) the proposed method enables information fusion among different data sources and jointly contributes to estimating the target 3D parameters.

## 2  Method

In this section, we first introduce the 3D Morphable Model, followed by our network architecture to obtain the DAFI. The Convolutional Point-set Representation (CPR) is described next. Under this framework, we introduce the consistency-aware parameter updating and CPR fusion approaches.

**3D Morphable Model** In the following, we introduce the base 3D model employed in this paper. The matrix $\mathbf{X}$ represents the 3D Morphable Model (3DMM) as proposed by Blanz *et al.* [18], which is a linear combination of a mean shape $\bar{\mathbf{X}}$, identity bases $\mathbf{A}_{id}$, and expression bases $\mathbf{A}_{exp}$ as follows:

$$\mathbf{X} = \bar{\mathbf{X}} + \mathbf{A}_{id}\boldsymbol{\alpha}_{id} + \mathbf{A}_{exp}\boldsymbol{\alpha}_{exp}, \qquad (1)$$

where $\mathbf{A}_{id}$ and $\mathbf{A}_{exp}$ contain 199 and 29 bases, which come from the Basel 3D face model [19] and the FaceWarehouse expression model [20]. The vectors $\boldsymbol{\alpha}_{id}$ and $\boldsymbol{\alpha}_{exp}$ are the corresponding coefficients of identity and expression. A compact representation $\boldsymbol{\alpha} = [\boldsymbol{\alpha}_{id}, \boldsymbol{\alpha}_{exp}]$ is employed to summarize the deformation in 3DMM. To project the 3D model $\mathbf{X}$ onto a 2D image plane, the weak-perspective projection model is employed:

$$\mathbf{Y}^G = \mathbf{S}(s)\mathbf{R}_z(r_z)\mathbf{R}_y(r_y)\mathbf{R}_x(r_x)\mathbf{X} + \mathbf{t}_{x,y}, \qquad (2)$$

where $\mathbf{Y}^G$ denotes the projected 2D positions from $\mathbf{X}$, and $G$ indicates the 2D positions as generated by 3DMM. The matrix $\mathbf{S}$ is a $2 \times 3$ scaling matrix, with the elements $\mathbf{S}(1,1)$ and $\mathbf{S}(2,2)$ equal to $s$ while the others remain zero. The in-plane rotation matrix, $\mathbf{R}_z$, is determined by the roll angle of the head: $r_z$. Out-of-plane rotation matrix: $\mathbf{R}_x$ and $\mathbf{R}_y$, are determined by pitch $r_x$ and yaw $r_y$ angles of the head. The vector $\mathbf{t}_{x,y} = [t_x, t_y]$ denotes the offset of the model in a 2D plane, which is a translation vector. The objective of this paper is to estimate the values of the 3D deformation coefficients: $\boldsymbol{\alpha}$, as well as pose parameters: $\mathbf{p} = [r_x, r_y, r_z, s, t_x, t_y]$.

### 2.1  Generate densely annotated image

As shown in Fig. 2a, a modified U-Net is employed to translate the original image $\mathbf{I}$ into a DAFI (denoted as $\mathbf{D}$), with the RGB values corresponding to the normalized X, Y, and Z coordinates in the 3D model space. The readers are referred to [6,8] for more details about this type of DAFI. The proposed network performs two tasks simultaneously: (i) predicts the value of each pixel on the DAFI, (ii) predicts 1/0 in the foreground/background region. The network contains three groups of output layers, and each group outputs a down-sampled (or the original) DAFI and a corresponding binary mask to indicate the foreground and background regions. The network architecture is shown in Fig. 2a. We refer to this network as Heavily Supervised U-Net (HS-U-Net). The $L_1$ and cross-entropy losses are employed to supervise the network to estimate the DAFIs and masks in all groups. After obtaining the DAFI, the index of each non-zero pixel is retrieved from a generic 3D model based on the predicted RGB values as in [6]. To speed up the index retrieval, a KD-tree is constructed based on the X, Y, and Z coordinates of all vertices on the generic 3D model: $\bar{\mathbf{X}}$.

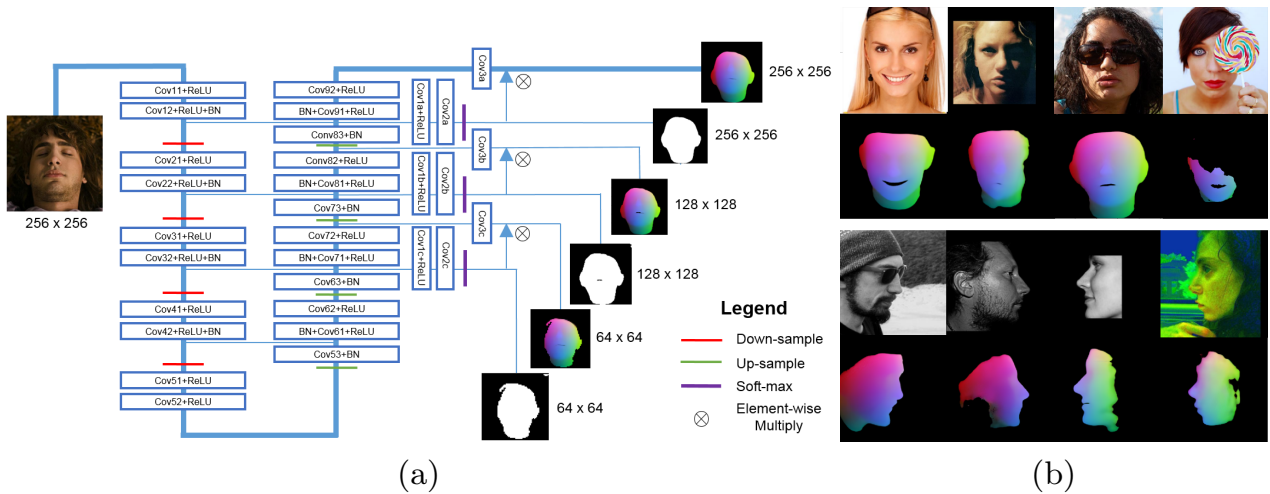

Fig. 2: (a) Depiction of the HS-U-Net employed for estimating DAFI. We employ three groups of loss layers to generate the final DAFI. (b) Depiction of the input images ($1^{st}$ and $3^{rd}$ rows) and the estimated DAFIs ($2^{nd}$ and $4^{th}$ rows). The degradation of DAFIs in challenging occlusions and illumination variations are presented.

The DAFI with the highest resolution is employed to construct the CPRs. In Fig. 2b, we present the estimated DAFIs under different poses, illuminations, and occlusions. Under the challenging scenario, DAFI contains more missing and mis-predicted pixels. A robust 3D face alignment approach is expected to extract useful information from the noisy data and learn to be robust to the missing and wrong values based on the contextual information encoded in the non-occluded pixels. Note that advanced image completion research [21] may help to alleviate the missing value problem. However, while the intuition of image completion is to use existing knowledge to infer the unknown, we use CPR to directly learn a robust mapping from DAFI to the estimated 3D parameters based on CPR.

### 2.2 Convolutional Point-set Representation (CPR)

Convolutional Point-set Representation employs matrices to represent the relations among one or multiple sets of annotated points in DAFI. It helps the convolution network to automatically encode non-linear relations among annotated pixels in a hierarchical manner, and contributes to estimating the parameters in 3D face alignment.

**Construct point-set relations with matrix:** If $\mathbf{Y}^O$ denotes the coordinates of the annotated pixels in DAFI, then CPR is derived to represent the relations between $\mathbf{Y}^O$ and $\mathbf{Y}^G$ (denoted in Eq. 2). To construct the CPR, $C$ vertices that have high probabilities to be localized accurately by U-Net among all the vertices of 3DMM are selected (more details about selecting the vertices will be introduced later). Letting $\mathbf{y}_i$ be the projected coordinate of one of the selected vertices in the DAFI, then a kernel $\phi_{i,j}$ is used to formulate the relation between $\mathbf{y}_i$ and another $\mathbf{y}_j$. Similarly, a triplet relation among $\mathbf{y}_i$, $\mathbf{y}_j$, and $\mathbf{y}_k$ is denoted as $\phi_{i,j,k}$. If vertex $i$ is not observable on DAFI, the values of all the related kernels are set to zero. A single channel CPR that describes the pixel-wise relations among $N$ out of $C$ vertices is formulated as:

$$\Lambda = \begin{bmatrix} \phi_{11} & \phi_{12} & \phi_{13} & \cdots & \phi_{1n} \\ \phi_{21} & \phi_{22} & \phi_{23} & \cdots & \phi_{2n} \\ \cdots & \cdots & \cdots & \cdots & \cdots \\ \phi_{n1} & \phi_{n2} & \phi_{n3} & \cdots & \phi_{nn} \end{bmatrix} \quad (3)$$

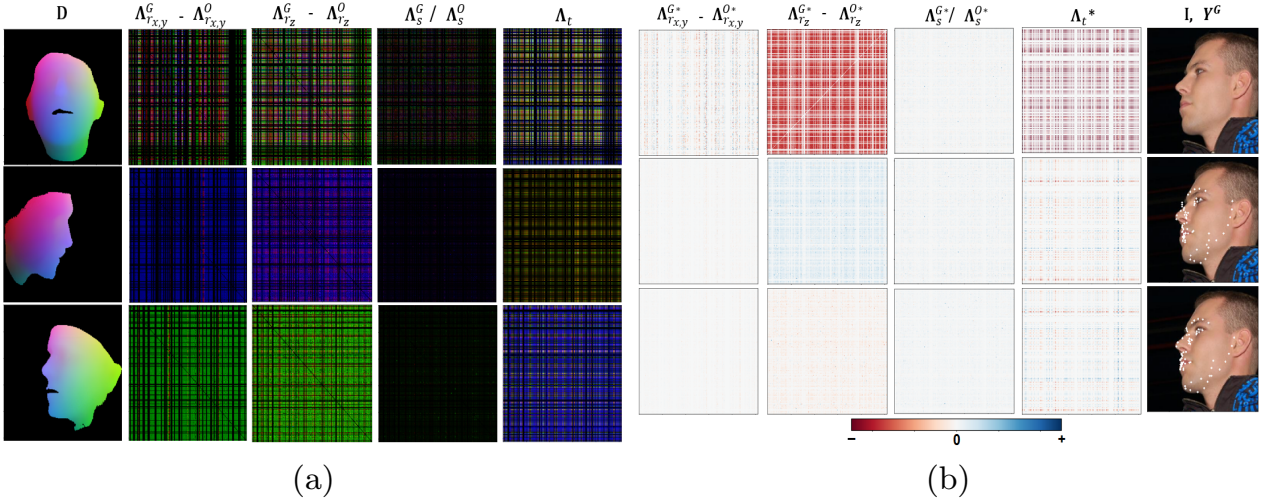

Fig. 3: (a) Depiction of the patterns of CPRs under different head pose variations. The $\Lambda^G$s are generated on a generic model $\bar{\mathbf{X}}$ with $\mathbf{p}^0$. (b) Depiction of the $2^{nd}$ channel of CPRs in the two rounds parameter updating (Fig. 1d). The color fades as the reprojection error goes smaller. The $1^{st}$ row corresponds to the initial status of CPRs. The $2^{st}$ row depicts the CPRs after the first round parameter updating. The $3^{rd}$ row depicts the CPRs after the second round parameter updating. The 68 landmarks on the projected 3D model are shown on the right. Note that for $\Lambda_t$, only the pattern along the x-axis is shown.

More channels (matrices) can be employed to represent all the relations in $C$ vertices if $N<C$. The volume made by one or multiple channels of $\Lambda$ is named convolutional point-set representation (CPR). For 3D face alignment, since minimizing the reprojection error between $\mathbf{Y}^O$ and $\mathbf{Y}^G$ is the target, CPR is employed as an intermediate representation in the optimization. This framework is depicted in Fig. 1c, where $\Lambda^G$ denotes the CPR generated from $\mathbf{Y}^G$ and $\Lambda^O$ denotes the CPR generated from $\mathbf{Y}^O$. The observed and estimated 2D coordinates of vertex $i$ in $\mathbf{Y}^O$ and $\mathbf{Y}^G$ is denoted as $\mathbf{y}_i^O$ and $\mathbf{y}_{i'}^G$ respectively.

**CPR as translation representation:** To estimate $\mathbf{t}$, the kernel in Eq. 3 is specified as: $\phi_{i,i'}^t = \mathbf{y}_i^O - \mathbf{y}_{i'}^G$. The CPR made by $\phi_{i,i'}^t$ is denoted by $\Lambda_{t_x}$ and $\boldsymbol{\Lambda}_{t_y}$ in 2D space. We use $\boldsymbol{\Lambda}_t$ to refer to $\boldsymbol{\Lambda}_{t_x}$ and $\boldsymbol{\Lambda}_{t_y}$ together. Because $\mathbf{y}^G_{i'}$ is generated from $\mathbf{Y}^G$, it is obvious that $\boldsymbol{\Lambda}_t$ is sensitive to the variations of $\mathbf{t}$, but also sensitive to the change of other pose parameters including $\mathbf{r}_x$, $\mathbf{r}_y$, $\mathbf{r}_z$, $s$ and $\boldsymbol{\alpha}$. So in order to determine $\mathbf{t}$ correctly, all the other parameters need to be estimated earlier.

**CPR as scaling representation:** To estimate $\mathbf{s}$, the contribution of $\mathbf{t}$ can be untangled. The distances among the vertices inside one of $\mathbf{Y}^O$ or $\mathbf{Y}^G$ are employed as kernel $\phi^s_{i,j} = ||\mathbf{y}_i - \mathbf{y}_j||$. The corresponding CPR is denoted as $\boldsymbol{\Lambda}_s$, which is in-variant of $\mathbf{t}$, but is sensitive to the changes in $\mathbf{r}_x$, $\mathbf{r}_y$, $\mathbf{r}_z$ and $\boldsymbol{\alpha}$. Thus, these parameters need to be estimated earlier.

**CPR as in-plane rotation representation:** To estimate $r_z$, the impacts of $\mathbf{t}$ and $\mathbf{s}$ can be untangled. The sine of the angle between two vectors is employed as kernel $\phi^{r_z}_{i,j} = \frac{||\mathbf{y}_i \times \mathbf{y}_j||}{||\mathbf{y}_i||||\mathbf{y}_j||}$ (Note that we do not employ the cosine angle because it cannot discriminate clockwise or counterclockwise rotation). The corresponding CPR is denoted as $\boldsymbol{\Lambda}_{r_z}$, which is in-variant of $\mathbf{t}$ and $\mathbf{s}$, but it is sensitive to the rotation parameters $\mathbf{r}_x$, $\mathbf{r}_y$, and $\boldsymbol{\alpha}$. These parameters need to be estimated earlier.

**CPR as out-of-plane rotation representation:** To estimate $r_x$ and $r_y$, it is important to untangle the impacts of $r_z$, $\mathbf{t}$, and $\mathbf{s}$ to the input space, so the cosine angle (dot product) between two unit vectors is employed as kernel $\phi^{r_{xy}}_{i,j,k} = \hat{\phi}_{i,j} \cdot \hat{\phi}_{j,k}$, where $\hat{\phi}_{i,j} = \frac{\mathbf{y}_i - \mathbf{y}_j}{||\mathbf{y}_i - \mathbf{y}_j||}$. The corresponding CPR is denoted as $\boldsymbol{\Lambda}_{r_{xy}}$, which is invariant to all the variations of $\mathbf{t}$, $\mathbf{s}$, but $\boldsymbol{\Lambda}_{r_{xy}}$ is still sensitive to changes in $\boldsymbol{\alpha}$.

**CPR as shape and expression representation:** To disentangle $\boldsymbol{\alpha}$ from $r_x$ and $r_y$, the depth image should be available, which means $\mathbf{y}_i$ has to be a 3D vector. The kernel $\phi^{r_{xy}}_{i,j,k}$ computed on the 3D $\mathbf{y}_i$, $\mathbf{y}_j$, and $\mathbf{y}_z$ can be directly employed to estimate $\boldsymbol{\alpha}$. The corresponding CPR is denoted as $\boldsymbol{\Lambda}_\alpha$. However, in our application, since the depth information of $\mathbf{y}_i$ is not available in DAFI, $\boldsymbol{\alpha}$ cannot be disentangled from $r_x$, $r_y$. Here, $\boldsymbol{\Lambda}_{r_{xy}}$ is employed as the CPR for $\boldsymbol{\alpha}$.

**Point-set selection for CPR:** Since CPR is designed as the input to a convolutional network, the memory limitation of convolutional networks requires a subset of 3D vertices to be selected to represent the point-set. The CPR employed in this paper contains three channels, which encode the relations of $C$ vertices. Since the human face can be considered symmetric, the first and second channel of CPR samples $C/2$ vertices from each side of the face. The third channel contains $C/2$ vertices sampled from both sides of the face, which are evenly selected from the vertices sampled in the first two channels. The vertices are randomly shuffled to minimize any local correlations.

The criteria for selecting $C$ are based on two probabilities, $p_1$ and $p_2$, which are computed on the estimated DAFIs on a reference database. Probability $p_1$ indicates the visibility of the landmark in different head pose variations. It takes into account how many times the vertex is visible in the reference database. For example, vertices on the nose have higher $p_1$ than the vertices under the jaw. Probability $p_2$ reflects the relative localization accuracy on DAFI, which is computed by measuring the distance between the estimated 2D positions on DAFI and the ground-truth projected positions on the reference. All the vertices are ranked based on the two probabilities, the top $C$ highest ranks vertices are picked to build CPR. Note that this is a preliminary way to select vertices for CPR, and we anticipate future research will explore more discriminative approaches.

**Visualization of CPR:** In Fig. 3a, multiple CPRs under different pose variations are depicted. Because we employ a 3-channel CPR, it can be visualized as an RGB image. Under large head pose variations, we note that chromatic CPRs appear mono, indicating that almost all elements in one channel of the CPRs are zeros.

### 2.3 Estimate Model Parameters with Deep Neural Networks

After introducing the CPR, it is important to describe how to use CPR to update the pose and shape parameters. Let CPRs extracted from the DAFI be denoted as $\boldsymbol{\Lambda}_s^O$, $\boldsymbol{\Lambda}_{r_z}^O$, and $\boldsymbol{\Lambda}_{r_{xy}}^O$. Similarly, the CPRs generated from $\mathbf{Y}^G$ are denoted as $\boldsymbol{\Lambda}_s^G$, $\boldsymbol{\Lambda}_{r_z}^G$, and $\boldsymbol{\Lambda}_{r_{xy}}^G$. The CPR for translation is denoted by $\boldsymbol{\Lambda}_t$. The distance matrix between a pair of a specific type of CPRs is computed and used to indicate the distance between the observed and the estimated model parameters. Then, the distance matrices are fed into convolutional networks followed by recurrent layers as in Eq. (4). The network architecture is shown in Fig. 4b.

The pose parameters $\mathbf{p}^0 = [r_x^0, r_y^0, r_z^0, s^0, t_x^0, t_y^0]$ are initialized as $[0,0,0,1,0,0]$. The $\boldsymbol{\alpha}$ is initialized to be all zero. The initial CPR generated based on $\mathbf{p}^0$ and $\boldsymbol{\alpha}^0$ is denoted as $\boldsymbol{\Lambda}_{r_{xy}}^{G_0}$. Then, the pose parameters are updated **sequentially** based on the transformation order in the weak-perspective projection model as follows:

$$\begin{aligned}
\alpha^q &= \alpha^{q-1} + \Phi_{RNN_0}(\boldsymbol{\Lambda}_{r_{xy}}^{G_{q-1}} - \boldsymbol{\Lambda}_{r_{xy}}^O; \mathbf{h}_\alpha^{q-1}) \\
[r_x^q, r_y^q] &= [r_x^{q-1}, r_y^{q-1}] + \beta_{DNN_1}^{q-1} \Phi_{RNN_1}(\boldsymbol{\Lambda}_{r_{xy}}^{G_{q-1}} - \boldsymbol{\Lambda}_{r_{xy}}^O; \mathbf{h}_{r_{x,y}}^{q-1}) \\
r_z^q &= r_z^{q-1} + \beta_{DNN_2}^{q-1} \Phi_{RNN_2}(\boldsymbol{\Lambda}_{r_z}^{G_{q-1}} - \boldsymbol{\Lambda}_{r_z}^O; \mathbf{h}_{r_z}^{q-1}) \\
s^q &= s^{q-1} + \beta_{DNN_3}^{q-1} \Phi_{RNN_3}(\boldsymbol{\Lambda}_s^{G_{q-1}}/\boldsymbol{\Lambda}_s^O; \mathbf{h}_s^{q-1}) \\
[t_x^q, t_y^q] &= [t_x^{q-1}, t_y^{q-1}] + \beta_{DNN_4}^{q-1} \Phi_{RNN_4}(\boldsymbol{\Lambda}_t^{q-1}; \mathbf{h}_{t_{x,y}}^{q-1}),
\end{aligned} \quad (4)$$

where $q$ indicates the index of stage in recurrent neural networks (RNN). The term $\mathbf{h}$ denotes the hidden parameter in RNN and $\beta_{DNN}^{q-1}$ denotes an adaptive weight which will be introduced in the next sub-section. The work flow depicted in Fig. 1d, contains two rounds. In the round-one, all the parameters are updated with $\beta = 1$ (the reason will be introduced next). In the second round, we employ consistency-aware parameter updating, which takes into account the reprojection error in updating. Note that the parameter $\boldsymbol{\alpha}$ only gets updated in round 1.

### 2.4 Consistency-aware parameter updating

In previous solutions of deep learning based 3D face alignment, the method terminated after applying the learned regressors to update the target pose/shape parameters. However, the method is unaware of whether the target reprojection error has been successfully minimized or not. The inclusion of DAFI allows for an extra feedback loop to inspect the change of reprojection error during parameter updating. The intuition for consistency-aware parameter updating is to use a weight to control the speed of parameter updating by observing the change of reprojection error along with a predicted direction. Intrinsically, this method is not constrained by the sampled vertices encoded in CPR, but able to use the "wisdom of the crowd" to passively check the global reprojection error via all the visible vertices in DAFI. This introduces addition robustness to the system since any unrealistic parameter updating will end up enlarging the reprojection error, which in turn degrades its own weight.

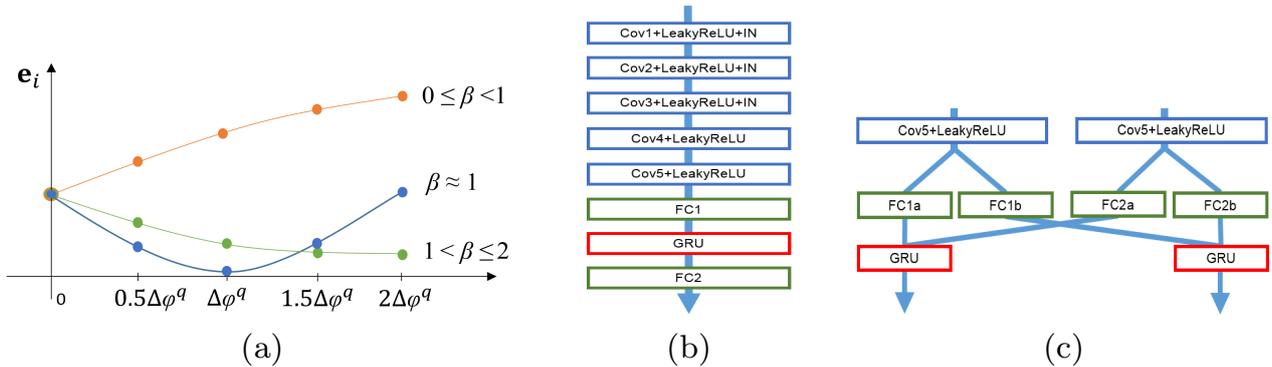

Fig. 4: (a) Depiction of the intuition for consistency-aware parameter updating. Weight $\beta$ is determined by the reprojection errors along with a sampling path, where the sampling number $M$ equals four in this example. Three lines are plotted to indicate the variations of reprojection error along the predicted updating direction. The orange line indicates the error is decreasing along the updating direction, which ldeas to a $\beta$ is smaller than 1. The green line indicates the error is enlarging along the updating direction, which end-up with a $\beta$ larger than 1. The blue line indicates estimated updating direction will end-up in a local minimum in terms of reprojection loss, leads to a $\beta$ close to 1. (b) Depiction of the convolutional network (with recurrent layer) employed for CPR regression. (c) Depiction of the fusing layers designed for merging information from a pair of CPRs.

The parameter updating of pose parameters depicted in Eq. 4 can be generally represented using the following equations:

$$\begin{aligned}\Delta\varphi^q &= \Phi_{RNN}(\boldsymbol{\Lambda}^G; \boldsymbol{\Lambda}^O; h^{q-1}) \\ \varphi^q &= \varphi^{q-1} + \beta^{q-1}\Delta\varphi^{q-1},\end{aligned} \quad (5)$$

where $\varphi^q$ represents any pose parameter(s) under updating. The term $\beta^{q-1}$ denotes a weight computed based on the reprojection error between $\mathbf{Y}^O$ and $\mathbf{Y}^G$ along a sampling path determined by $\Delta\varphi^q$. Specifically, a path along the direction of updating is derived as follows:

$$Path = \{\frac{\Delta\varphi^q}{M}, \frac{2\Delta\varphi^q}{M}, ... \frac{(2M-1)\Delta\varphi^q}{M}, 2\Delta\varphi^q\}, \quad (6)$$

Each sampling point on the *Path* corresponds to a different updating of the pose parameter, which will generate a different 2D projection $\mathbf{Y}_i^G$, where $i \in \{1 : 2M\}$. The reprojection error of each $\mathbf{Y}_i^G$ can be calculated as: $e_i = ||\mathbf{Y}_i^G - \mathbf{Y}^O||$. The number of vertices employed for computing this error is set to be $\kappa$ out of the total detected vertices in DAFI. The weight $\beta$ in Eq. 4 can be learned by a deep neural network with few hidden layers:

$$\beta = \Phi_{DNN}([e_1, e_2...e_{2M}]), \tag{7}$$

where the vector $[e_1, e_2...e_{2M}]$ contains reprojection errors for each sampling point. The result $\beta$ is forced to be in a range between 0 and 2. It is computed based on the pattern of reprojection errors and determined by whether the $\Delta\varphi^q$ can bring the reprojection error into a local minimum along a predicted direction. The intuition of Eq. 6 and Eq. 7 can be visualized in Fig. 4a.

Look back into Eq. 4, the DNN and RNN are learned jointly in training. Note that the reprojection errors in Eq. 7 are only meaningful after all the parameters have been updated at least once, so $\beta$ is forced to be 1 in the first round of parameter updating. Correspondingly, Fig. 3b clearly visualizes the pattern variations in the second channel of CPRs under the two rounds of parameter updating.

### 2.5   Fusing Multiple Convolutional Point-set Representations

Note that CPR is a general way of representing the relations of annotated pixels in a 2D image. It can be employed to encode the set of landmarks generated by state-of-the-art landmark detectors. The advantage of using an automatic landmark detector is that it can provide limited but highly robust landmark annotations on the most salient points on the face. This has the potential to provide reliable regularizations for 3D face alignment.

To merge the sensitivity of CPR-$C_1$ that is constructed based on DAFI, and the reliability of CPR-$C_2$ that is constructed based on sparse landmarks, we fuse the sixth layer of the CPR network (Fig. 4b) through a crossing architecture as depicted in Fig. 4c. The merged model performs better than a single model. The 3D parameters are then updated by the following:

$$\varphi^q = \varphi^{q-1} + \beta_{C_1}^{q-1} \Delta\varphi_{C_1}^{q-1} + \beta_{C_2}^{q-1} \Delta\varphi_{C_2}^{q-1}, \tag{8}$$

where $[\beta_{C_1}^{q-1}, \Delta\varphi_{C_1}^{q-1}]$ and $[\beta_{C_2}^{q-1}, \Delta\varphi_{C_2}^{q-1}]$ are predicted from CPR-$C_1$ and CPR-$C_2$ respectively.

## 3   Experiments

### 3.1   Datasets

**The 300W database** [8] is employed for training the deep networks in this paper. The database contains $122,450$ rendered facial images which cover large pose spectrum (up to $90°$ yaw variations). Before training, about 1% of images are reserved as a reference database to select the discriminative pixels for CPR.

**The AFLW2000 database** [8] is an "in-the-wild" database. It contains the first 2,000 images of AFLW [22], for each one has the ground-truth shape and pose parameters, making it a good database for evaluating the performance of 3D face alignment. The reprojection error of 68 landmarks is reported on this database.

**The AFLW database** [22] is an "in-the-wild" database. It contains large head pose variations. It is employed for evaluating the 2D landmark localization error based on 3D face alignment. As in [8], 21,080 images are employed to evaluate the algorithm's performance. The reprojection error of 21 landmarks is reported on this database.

**The COFW database** [23] is an "in-the-wild" database. It contains images depicting severe facial occlusion. We use the 507 testing images to evaluate the robustness of 3D face alignment approaches. The reprojeciton error is reported on the 51 inner landmarks out of 68 landmarks annotated by Ghiasi *et al.*[24].

### 3.2  Implementation details

The HS-U-Net is trained on the 300W-LP database for 40 epochs from scratch. The size of the input image is $256 \times 256$. The database is augmented by rotation, translation, and scaling operations. ADAM optimizer [25] is employed for updating the network parameters. The learning rate is gradually decreased from $10^{-4}$ to $10^{-7}$. The batch size is set to be four. The trained network is evaluated on the reference database, and $C = 512$ vertices are selected for CPRs. The dimension of CPRs is set to be $256 \times 256$. As for parameter updating, the number of the recurrent stage ($q$) is set to be three. The number of sampling points for consistency-aware parameter updating is set to be 10 ($M = 5$). The sampling ratio $\kappa$ for computing the reprojection error is set to be 0.1. The recurrent networks for updating parameters take one epoch to converge (from scratch). We trained a different set of networks for consistency-aware parameter updating. It also takes one epoch to converge (from scratch). The batch size is set to be one when training all the network for parameter updating, and the learning rate is set to be $10^{-4}$. In application, we observed that the RNN for estimating the translation parameter produces little improvement over direct estimation it through CPR, so we compute the mean of non-zero values on $\boldsymbol{\Lambda}_{t_x}$ and $\boldsymbol{\Lambda}_{t_y}$ to be $t_x$ and $t_y$. Normalized Mean Errors (NME) are reported in all the experiments as in the previous paper [8].

### 3.3  Accuracy of Densely Annotated Facial Image

The objective of this experiment is to comprehensibly evaluate the accuracy of DAFI in terms of vertex annotation error. This has not yet been studied in previous research [6,9] that relies on DAFI. The results are shown in Table 1. FAN-3D is selected to be the baseline. We report the normalized mean error (NME) of annotating each specific vertex on 3DMM along with the average detection rate (ADR) of the vertex. An annotated vertex (pixel) is defined to be 'detected' if its RGB value on DAFI can be successfully matched to a vertex in the generic 3D model with an Euclidean distance smaller than 0.01. For example, if 50 out of 512 vertices are detected on DAFI, the detection rate is computed as: 50/512 = 0.098 <10%. The FAN-3D is selected to be the baseline. Since is a heatmap based landmark detector, it is guaranteed that all 68 facial vertices can be detected. Because FAN-3D also estimates the locations of occluded landmarks, the NME is larger than the DAFI-based approach.

Table 1: Depiction of the NME (%)/ADR (%) of vertex detection on 2,000 images of AFLW-3D database under different yaw variations. 'Failure' indicates the percentage of samples whose NME > 10% or vertex detection rate < 10%. The (Num. of vert.) represents the number of vertices involved in computing the annotation error on each image. The 3DMM contains 53215 vertices in total.

| Method (Num. of vert.) | AFLW2000-3D Dataset | | | | | |
|---|---|---|---|---|---|---|
| | [0°,30°] | [30°,60°] | [60°,90°] | Mean | Std | Failure |
| FAN-3D (68) | 2.67/**100** | 3.76/**100** | 5.24/**100** | 3.89/**100** | 1.05/**0** | **1.25** |
| DAFI (68) | 2.35/37.4 | 2.84/28.1 | 3.75/17.7 | 2.98/27.7 | 0.58/8.05 | 2.75 |
| DAFI (53215) | 2.63/29.0 | 3.11/27.5 | 4.01/22.7 | 3.25/26.4 | 0.57/2.69 | 2.05 |
| DAFI (512 for CPR) | **2.17**/49.3 | **2.61**/42.1 | **3.50**/32.2 | **2.76**/41.2 | **0.55**/7.01 | 1.60 |

It is observed that the selected 512 vertices for CPR have higher ADR and lower NME than the 68 hand selected vertices. This shows that once the selected vertices are detected on DAFI, they have smaller NME than FAN-3D. This observation demonstrates that DAFI generated by HS-U-Net can be used as a good basement for 3D face alignment. The result also indicates that the challenge for 3D alignment lies in vertex detection. It is not certain that whether a specific vertex is available on a DAFI and how it contributes to estimating pose/shape parameters. Due to the pattern of occlusion encodes the pose-specific information, our goal is to let all the detected and occluded (missing) vertices serve to minimize the overall reprojection error through CPRs.

### 3.4  Ablation Study

The objective of this experiment is to systematically analyze the contributions of different proposed modules in CPR-based 3D face alignment. The reprojection errors (2D) of 68 landmarks are reported, and the CED curve is available in Fig. 5a. The method of '3DDA+SDM' [8] is selected to be the baseline. In this experiment, 'PnP-FAN3D-68' refers to the classical method for 3D face alignment, widely employed for face analysis [26, 27]. First, a state-of-the-art landmark detector like FAN3D is used to detect landmarks. Then, the perspective-N-point (PnP) method (implemented by OpenCV [28]) is used for 3D face alignment. In 'PnP + DAFI(512)', all the visible vertices among the 512 selected vertices are employed as input to PnP for 3D face alignment (if DAFI detects no vertice on the face, the NME is set to be 1). A similar approach has been used in [6]. The results demonstrate that the vertices annotated by DAFI improved face alignment. In 'CPR(512)', the CPRs generated by 512 selected vertices are employed for 3D face alignment. It is a single round approach without using Consistency Aware Parameter Updating (CAPU). The 'CPR-H' refers to the approach that employs the CPR-fusion method introduced in Sec. 2.5, which merged the infor-mation from the 68 landmarks generated by FAN3D. It can be observed from the CED curve that the method handles some extreme cases when the CPR(512) fails. The 'CPR(H)+CAPU' refers to the two-round approach that using CAPU introduced in Sec. 2.5 to estimate parameters. The CAPU further improved the performance of 3D face alignment. In the following section, this approach is abbreviated as CPRFA (CPR-based face alignment).

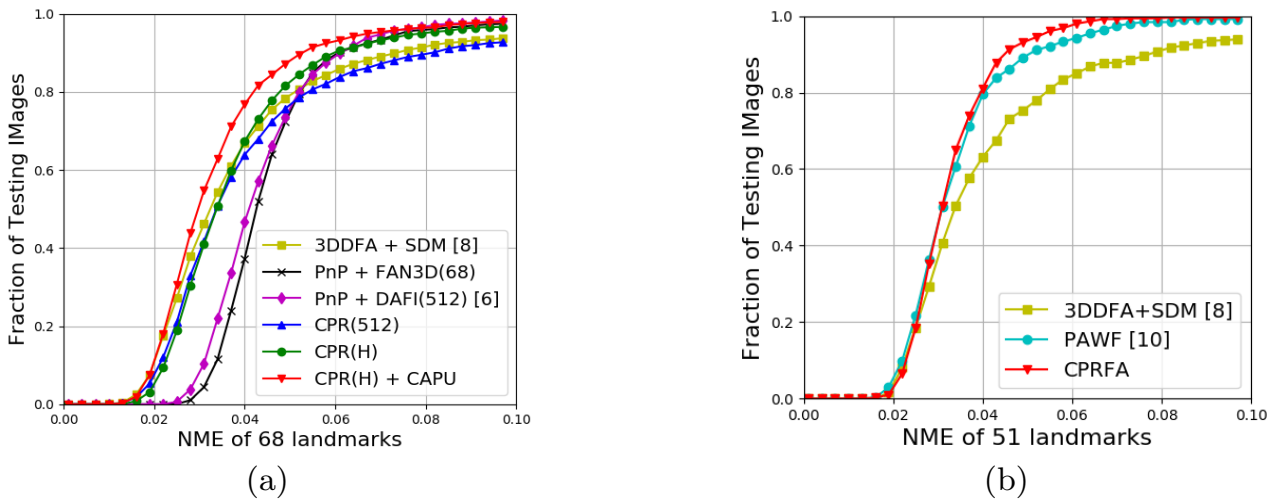

Fig. 5: Depiction of the CED curves of NME (%) on (a) AFLW2000 and (b) COFW database.

### 3.5  Comparison Experiments

The objective of this experiment is to compare CPRFA with state-of-the-art approaches that using a 3D model for 3D face alignment. Under this objective, 3DDFA[8], LDFC [9], 3DSTN [12], and PAWF [10] are selected to be the baselines, as they are the most recent state-of-the-art for 3D face alignment. Note that the face alignment of LDFC is also based on DAFIs, is the closest baseline to ours. The performance of the selected methods is reported in Table 2 and Fig. 6.

**Results on AFLW2000 and AFLW databases** indicate that CPRFA significantly outperforms LDFC. We believe this is due to two reasons: First, in contrast to LDFC, generating DAFI by HS-U-Net does not rely on a template image. So it is more robust to special cases where the input image is significantly different from the template image. Second, in CPRFA, deep neural networks are employed to estimate the model parameters based on learned mappings from CPRs, so it is robust to missing and mistaken predictions on DAFI, while LDFC employs the less robust on-line optimization approach to tackle this problem. It is observed from the Table 1 that our model achieves the best face alignment result on the AFLW2000 database and has the minimum standard deviation on both databases, which demonstrates that it is a good candidate for 3DMM-based face alignment. We also notice that 3DSTN employs a Thin Plate Spline (TPS) model for face alignment, is more flexible than the 3DMM that the other three methods (3DDFA, LDFC, and CPRFA) employed in terms of deforming itself to fit the 21 2D landmarks on the AFLW database. This helps to explain

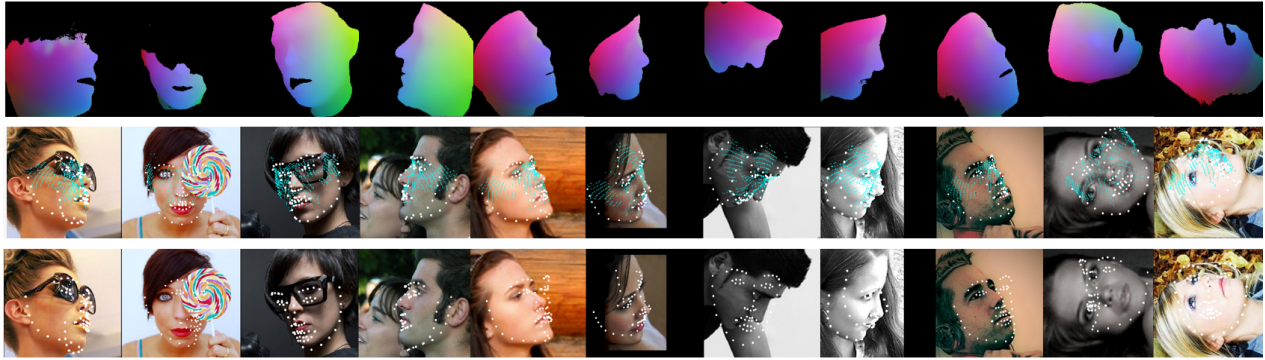

Fig. 6: Depiction of alignment results with failure cases. (T) Estimated DAFI (M) Results of CPRFA, with blue dots depicting the projected positions of the 512 selected vertices to construct the CPRs. (B) 3D alignment results of 3DDFA + SDM.

Table 2: The NME (%) of face alignment on AFLW and AFLW2000 database.

| Method | AFLW Database (21 pts) | | | | | AFLW2000-3D Database (68 pts) | | | | |
|---|---|---|---|---|---|---|---|---|---|---|
| | [0°,30°] | (30°,60°] | (60°,90°] | Mean | Std | [0°,30°] | (30°,60°] | (60°,90°] | Mean | Std |
| 3DDFA | 5.00 | 5.06 | 6.74 | 5.60 | 0.99 | 3.78 | 4.54 | 7.93 | 5.42 | 2.21 |
| 3DSTN | **3.55** | **3.92** | **5.21** | **4.23** | 0.87 | 3.15 | 4.33 | 5.98 | 4.49 | 1.42 |
| LDFC | 5.94 | 6.48 | 7.96 | 6.79 | 0.85 | 3.62 | 6.06 | 9.56 | 6.41 | 2.44 |
| **CPRFA** | 3.71 | 4.20 | 5.52 | 4.47 | **0.76** | **3.12** | **3.62** | **5.40** | **4.05** | **1.20** |

why 3DSTN is the only method in Table 2 that performs better on AFLW than AFLW2000-3D.

**Results on COFW database:** The CDE curve is plotted in Fig. 5b. ThIS experiment further demonstrates that CPRFA is more robust to face occlusions than the other 3D face alignment approaches.

**Running speed:** CPRFA takes 0.9s to process one image on a single Nvidia GTX 1080Ti GPU. Estimating DAFI takes 0.03s, the first round of parameter updating takes 0.4s, the second round takes 0.3s. Applying FAN3D for CPR-fusing takes 0.12s, other initial computation takes 0.05s. The code is completely implemented in Python with Pytorch [29]. Currently, our implementation is slower than 3DDFA (C++, 75ms), 3DSTN (19ms), but faster than PAWF (5s).

## 4   Conclusions

We proposed a convolutional bridge to enable convolutional neural networks to distill high-level knowledge for 3D face alignment from a densely annotated facial image (DAFI). The method is demonstrated to be robust to missing and mistaken values in DAFI, and outperforms previous solutions that use DAFI or deep cascaded regression. We proposed two modules for this framework to update model parameters. The modules employ reprojection error and information fusing techniques to further boost the performance. The proposed approach can be applied to 3D facial analysis in-the-wild and has the potential to be generalized for solving object pose estimation and deformation problems for which a densely annotated image is available.